\documentclass{article}

\pdfoutput=1
\usepackage[accepted]{icml2025}

\usepackage{graphicx}
\DeclareGraphicsExtensions{.pdf,.png,.jpg,.jpeg}
\usepackage{tipa}
\usepackage{subcaption}

\usepackage[utf8]{inputenc} 
\usepackage[T1]{fontenc}    
\usepackage{hyperref}       
\usepackage{url}            
\usepackage{booktabs}       
\usepackage{amsfonts}       
\usepackage{nicefrac}       
\usepackage{microtype}      
\usepackage{xcolor}         
\usepackage{amsmath} 

\usepackage[english]{babel}
\usepackage{CJKutf8}

\begin{document}

\twocolumn[
\icmltitle{I Have No Mouth, and I Must Rhyme: Uncovering Internal Phonetic Representations in LLaMA 3.2}



\icmlsetsymbol{equal}{*}

\begin{icmlauthorlist}
\icmlauthor{Oliver McLaughlin}{sch}
\icmlauthor{Arjun Khurana}{sch}
\icmlauthor{Jack Merullo}{sch}
\end{icmlauthorlist}

\icmlcorrespondingauthor{Oliver McLaughlin}{oliver\_mclaughlin@brown.edu}
\icmlcorrespondingauthor{Arjun Khurana}{arjun\_khurana@alumni.brown.edu}
\icmlcorrespondingauthor{Jack Merullo}{jack\_merullo@brown.edu}

\icmlaffiliation{sch}{Brown University, Providence, Rhode Island}

\icmlkeywords{Machine Learning, ICML}

\vskip 0.3in
]



\printAffiliationsAndNotice{}  

\begin{abstract}
Large language models demonstrate proficiency on phonetic tasks, such as rhyming, without explicit phonetic or auditory grounding.
In this work, we investigate how \verb|Llama-3.2-1B-Instruct| represents token-level phonetic information. Our results suggest that Llama uses a rich internal model of phonemes to complete phonetic tasks. We provide evidence for high-level organization of phoneme representations in its latent space. In doing so, we also identify a ``phoneme mover head" which promotes phonetic information during rhyming tasks. We visualize the output space of this head and find that, while notable differences exist, Llama learns a model of vowels similar to the standard IPA vowel chart for humans, despite receiving no direct supervision to do so.

\end{abstract}

\section{Introduction}
Language models (``LMs") cannot hear speech. In spite of this, many large LMs can consistently produce rhymes, poetry, alliteration, and other phenomena which seem to require a fundamental understanding of phonetic properties.  We propose that these models complete these tasks using underlying token-level phonetic representations as well as circuits to retrieve phonetic information.

Up to this point, study of LMs' phonetic behavior has been largely limited to training models on grounded phonetic information \cite{english23_interspeech, popescu-belis-etal-2023-gpoet}. Anthropic's recent work \textit{On the Biology of a Large Language Model} \cite{lindsey2025biology} uses attribution graphs to provide evidence of rhyme planning circuits in Claude 3.5 Haiku, but they do not rigorously study the internal phonetic representations that allow these circuits to function.

It is reasonable to ask the extent to which an LM is able to infer phonetic properties directly from tokens. Following results that LMs and even simple text models can extract information about the world without direct supervision \citep{louwerse2012representing, mikolov2013distributed, gurneelanguage}, we hypothesize that LMs also learn a robust, structured model of phonetic information that supports interventions.
We investigate how Llama 3.2 \cite{llama3} represents phonetic  information through methods commonly used in interpretability analysis and find evidence to support this hypothesis: rather than simply memorizing phonetic information for various tokens, Llama uses structure in its latent space across tokens to represent the phonetics of given input tokens.

Our experiments follow a straightforward methodology. To explore linear token-level phonetic information, we employ linear probes to identify subspaces of the residual stream and embedding and inspect the subspaces. To explore the mechanisms that leverage this information, we isolate components of the model which prove to be impactful in expressing the LM's phonetic beliefs. Using this simple approach, we find evidence of rich phonetic representations within \verb|Llama-3.2-1B-Instruct|.

\begin{figure}
    \centering
    \includegraphics[width=0.78\linewidth]{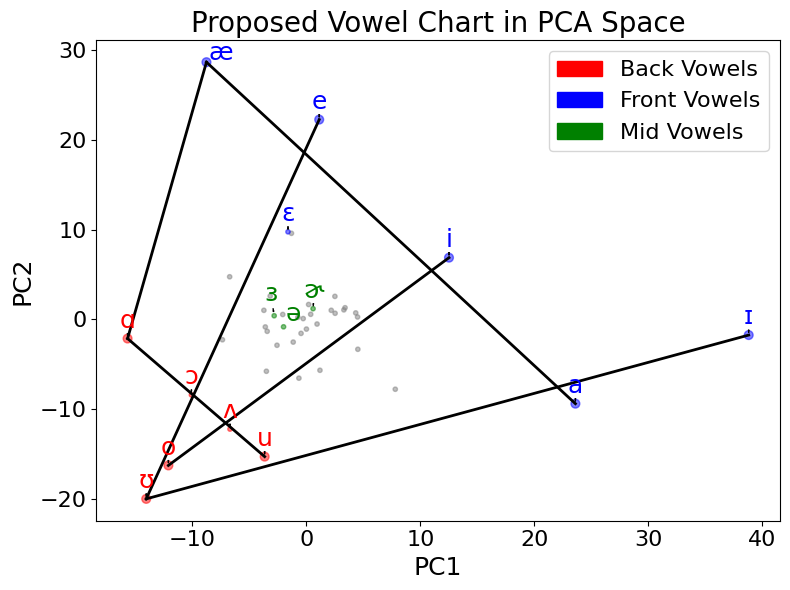}
    \caption{Patterns emerge from vowel representations
    under our methodology, revealing a world model partially inconsistent with human anatomy.
    }
    \label{fig:new_ipa_over_pca}
\end{figure}

In Section \ref{embeds_1}, we find vectors in the embedding space corresponding to common English phonemes. We perform causal interventions in the embedding space using these vectors to alter the model's performance on a rhyming task, demonstrating their role in rhyming processes. In Section \ref{phon_mover_head} we identify a single ``phoneme mover head" using activation patching. We decode this head's result vectors using logit lens \citep{nostalgebraist2020interpreting} and demonstrate that the phonetic information in this head is, to some extent, cross-lingual. In Section \ref{geometry}, we use the result vectors of this phoneme mover head to perform a phonetically-informed dimensionality reduction on the embedding space in order to analyze the geometry of the phoneme vectors from Section \ref{embeds_1}. We see consistent linear patterns across internal representations of vowel phonemes (Figure \ref{fig:new_ipa_over_pca}).

These patterns differ from anatomical properties of vowel phonemes, suggesting the presence of a robust internal vowel model which is distinct from human vowel models. We construct an organized representation of these patterns and contrast it with existing anatomically-grounded vowel organization systems.


\section{Phonetic Information in Token Embeddings} 
\label{embeds_1}

In Llama 3.2, like in most open-source LMs, the tokenization process does not explicitly encode any phonetic information about input tokens. We hypothesize that some phonetic information is encoded through token embeddings. 
To investigate this hypothesis, we train a multi-hot linear probe to predict which phonemes are present in a word from its embedding.
We differentiate phonemes based on the International Phonetic Alphabet, using WikiPron \cite{wikipron} as a pronunciation reference.

\paragraph{Probing the embedding space} 
\label{Probing_Embedding}
Our probe predicts the correct phonemes for approximately $96\%$ of single-token words, compared with $42\%$ for the same probe architecture trained on embeddings from a randomly-generated embedding matrix. This indicates that the model's embedding matrix encodes some amount of recoverable phonetic information.

Since this probe is linear, it essentially constitutes a linear map from Llama 3.2's 2048-dimensional embedding space to what we call an ``IPA phoneme space": a 44-dimensional space with one axis for each common English-language phoneme. Each row of our probe matrix, therefore, could be considered a representation of its corresponding phoneme in the embedding space.

\begin{figure}[htb]
    \centering
    \includegraphics[width=1.0\linewidth]
    {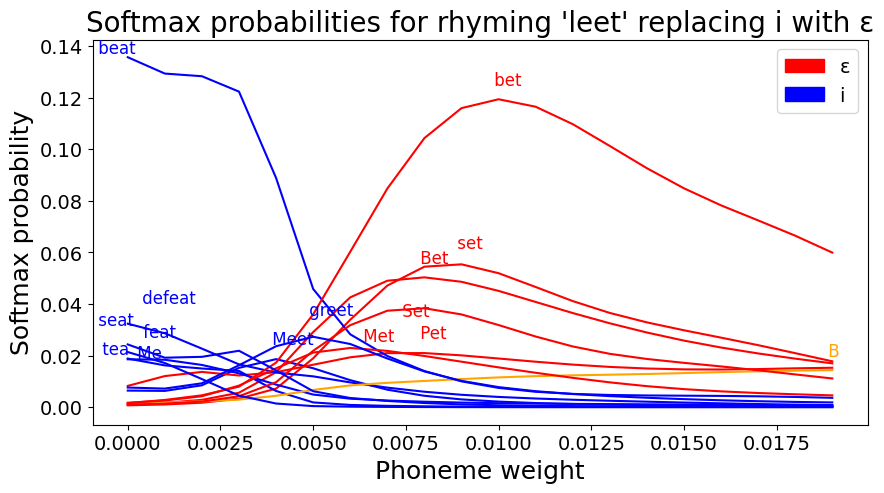}
    \caption{Example intervention on rhymes with \texttt{leet}. Blue indicates that the $\xi$ phoneme \textipa{/i/} is present, red indicates that the $\mu$ phoneme \textipa{/E/} is present.}
    \label{fig:leet}
\end{figure}

\paragraph{Causal interventions on embeddings}
To test the degree to which the rows of our probe represent phonemes in latent space, we run a causal intervention experiment to change the expected rhyming output for a given target word by intervening on embedding vectors. We test the model on a simple rhyming task:

\begin{verbatim}
prompt = """Here are a few examples of words
that rhyme with <word>:"""
\end{verbatim}

where \verb|<word>| is a single-token word with one unique vowel sound. We then select two ``phoneme vectors" (rows of our probe matrix) in the embedding space: $\xi$, the phoneme vector corresponding to the vowel in \verb|<word>|; and $\mu$, the phoneme vector corresponding to a different vowel. We perform a forward pass of Llama with the following intervention at the embedding step:
\[
E = E +c(\mu-\xi)
\]
where $E$ is the embedding vector corresponding to our rhyming word \verb|<word>|, and $c$ is a scalar. As we increase $c$ (the weight of our intervention), the model's prediction tends to switch from words with the $\xi$ vowel to words with the $\mu$ vowel. Figure \ref{fig:leet} shows example model output results where \verb|<word>| is \verb|leet| pronounced \textipa{/li:t/}, $\xi$ corresponds to \textipa{/i/}, and $\mu$ corresponds to \textipa{E}.

\section{Phoneme Mover Head}

\begin{figure}[htb]
    \centering
    \includegraphics[width=1.0\linewidth]{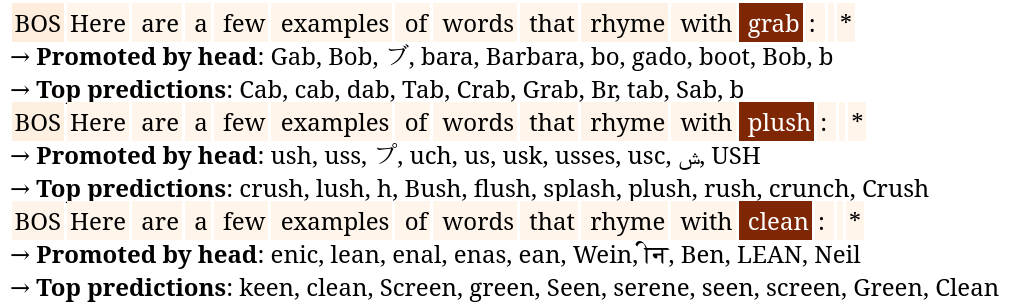}
    \caption{Across three examples: Attention patterns of H13L12 at the final token shows the head attends to the rhyme target token. Running logit lens on the corresponding result vector (shown after $\rightarrow$) produces phonetically similar tokens.}
    \label{fig:pattern}
\end{figure}

\label{phon_mover_head}
Given that embeddings seem to contain phonetic information, we seek to identify model components which use this information to complete rhyming tasks. We run activation patching experiments over all attention heads and MLP components across all layers.

\paragraph{Patching setup} \label{patching_setup}
To patch, we perform two full passes of the model using parallel prompts containing two sufficiently different\footnote{Two words are ``\textit{sufficiently different}" iff there is no third word that rhymes with both simultaneously.} rhyme target words. For example, setting \verb|<word>="clean"| in our above template results in the top predicted token \verb|"keen"|. Similarly, setting \verb|<word>="track"| results in \verb|"back"|. These then form our ``clean" and ``corrupted" runs (in line with \citet{meng2023locatingeditingfactualassociations}) allowing us to inspect the mean normalized logit difference between \verb|"keen"| and \verb|"back"|. Head 13 of Layer 12 (``H13L12") emerged as the most critical for our rhyming task with a mean normalized logit difference of 0.48 (out of 1). This was significantly higher than both the mean (0.002) and the second highest value (0.19). See Appendix \ref{patchingres} for a visual representation. 

\paragraph{Result vectors and cross-lingual features}
To understand the effect of this head on task completion, we explore its contributions to the residual stream by inspecting its result vectors\footnote{i.e. the projection of a single head back to the residual stream dimension through its output matrix, see \citet{elhage2021mathematical}}. We decode these vectors into the vocab space using logit lens \cite{nostalgebraist2020interpreting} in order to study the tokens this head promotes. Figure \ref{fig:pattern} demonstrates these results along with last-token attention patterns. We detail further study of the head dimension in Appendix \ref{Exploring}.

There is a clear, if approximate, phonetic similarity between the target rhyme word and the decoded result vectors of H13L12. These results and the associated attention patterns suggest that this head moves phonetic information from the target word to the final token's residual stream. Because of this apparent phonetic promotion behavior, we call this head a ``phoneme mover head". We investigate the connection between the result vectors produced by the head and completion of our rhyming task in Appendix \ref{resvec}.

We also see clear evidence of the LM modeling phonetic information across languages, as decoding result vectors produces language-agnostic sets of similar phonemes. In Figure \ref{fig:pattern}, for instance, we see that one of the top tokens promoted into the residual stream for \verb|plush| is ``sh (Arabic)" \textipa{/\textesh/}, for \verb|clean| we see ``een (Hindi)" (\textipa{/i:n/}), and for \verb|grab| we see ``bu (Japanese)" (\textipa{/b\textturnm/}), among others. 

\paragraph{Further Phonetic Heads}
Upon further inspection we discovered a set of heads (H13L12, H21L14, H22L14) with nearly identical attention patterns. After zero ablating all three of these heads, we noticed that the model could no longer correctly produce a \textit{single} rhyming token. Instead, it produced a first token (typically a single letter) and then a corresponding second token which typically completed the response word (For example if the target word was "plush" we might get "l" "ush"). Results were generally mixed as to if the response produced was truly a correct rhyme or not. Omitting any one of these three heads from ablation reinstated the model's original rhyming ability. Of further interest is that the composition scores \cite{elhage2021mathematical} for all three of these heads and all prior heads were also essentially identical, suggesting a common channel.

\section{Geometry of Phoneme Vectors}\label{geometry}
\begin{figure}[htb]
    \centering
    \includegraphics[width=0.7\linewidth]{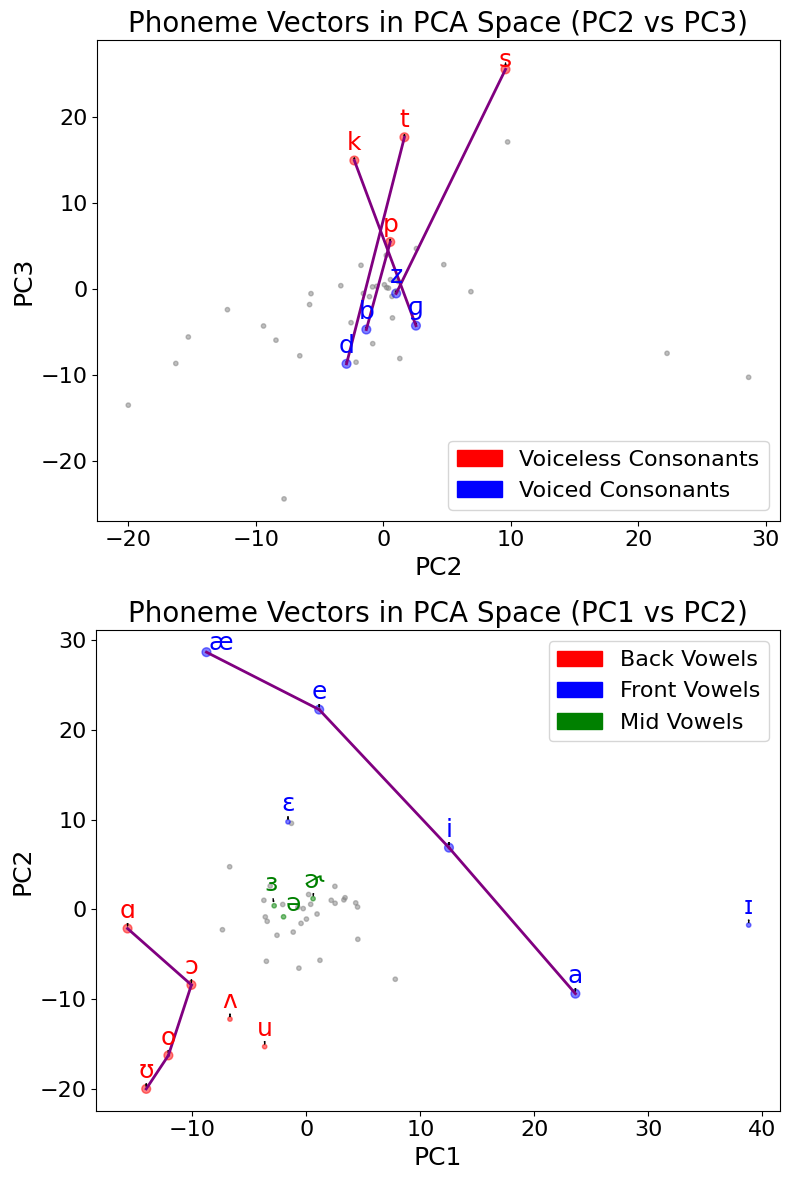}
    \caption{
    Phoneme vectors reduced to 2 dimensions using PCA trained on H13L12 result vectors. Phonetic patterns emerge among consonant voicedness (top) and vowel backness (bottom).
    }
    \label{fig:charts}
\end{figure}
We visualize the result vectors of Head H13L12 for 5742 different target words on our template with principal component analysis (PCA) to understand how the model organizes phonetic information.
\begin{figure*}[ht!]
    \centering
    \begin{subfigure}[t]{0.45\textwidth}
        \centering
        \includegraphics[width=.8\textwidth]{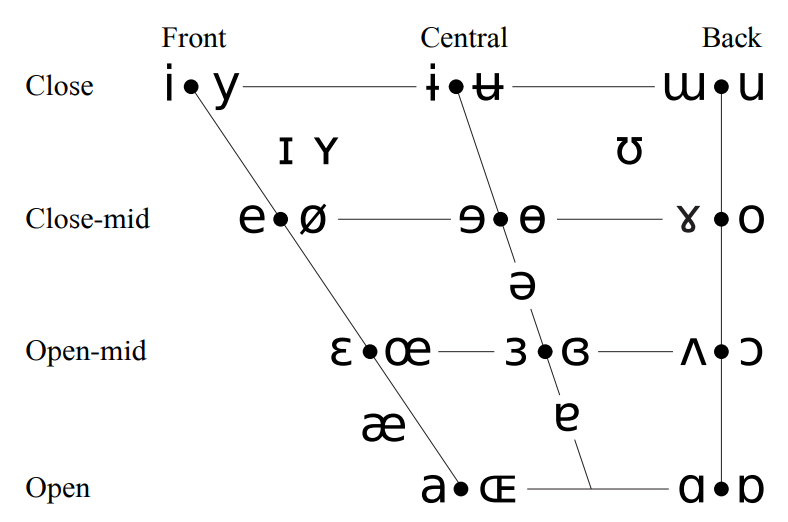}
    \end{subfigure}%
    \begin{subfigure}[t]{0.45\textwidth}
        \centering
        \includegraphics[width=\textwidth]{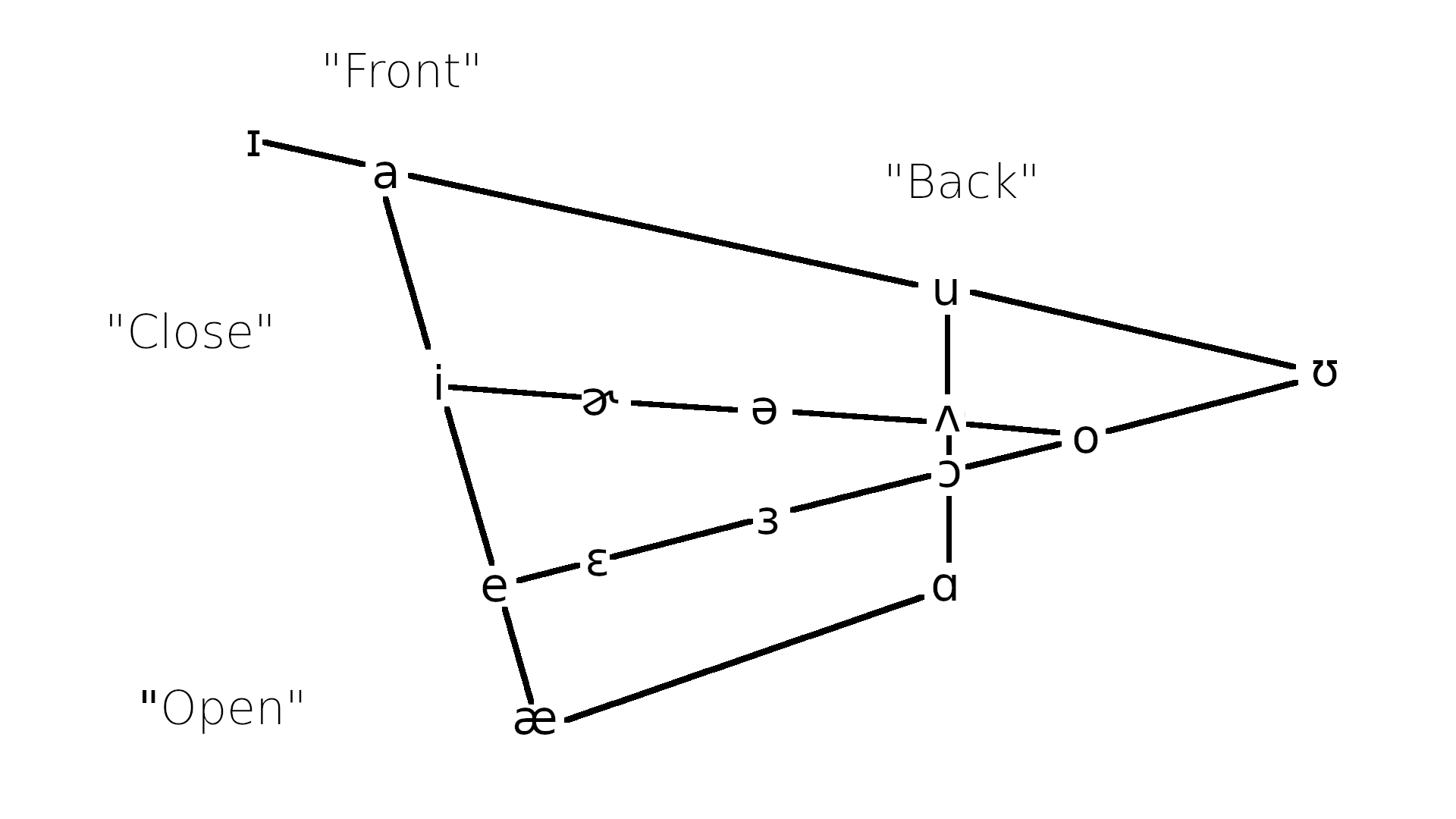}
    \end{subfigure}
    \caption{Standard IPA representation of vowels in humans (left) and a proposed IPA-style representation of Llama's internal vowel model (right). The positions of some phonemes, such as \textipa{/a/} and \textipa{/I/}, notably differ.}
    \label{fig:ipa_chart}
\end{figure*}
We expect that if the model has an organized representation of phonetic information, we should see similar phonemes cluster together regardless of their source word.

For consonants, Principal Components 2 and 3 contain the least noisy phonetic information, in line with findings from \citet{engels2025languagemodelfeaturesonedimensionally}. Figure \ref{fig:charts} shows that phoneme vectors corresponding to voiced consonants versus their voiceless equivalents follow a consistent linear pattern across PC3.

Vowels also follow consistent patterns in this space. These patterns partially align with 
anatomically-based IPA distinctions, but there are some key differences. Figure \ref{fig:charts} shows that PC1 and PC2 seem to distinguish vowel backness: front vowels tend to have positive PCs, mid vowels center around the origin, and back vowels tend to have negative PCs. Likewise, 
vowel openness order is consistent within backness classes:
more closed vowels have lesser PC2 components than more open vowels in the same backness class.

The notable exceptions are the extreme vowels. For instance, \textipa{/a/}, an open front vowel, has a large PC1 and small PC2, placing it at the ``close" end of the front vowels in PCA space.
\textipa{/I/}, a near-close near-front vowel, has an unusually large PC1 value for its backness. 

\paragraph{Llama's vowel chart}
It is important to remember that these phonetic representations are not explicitly grounded in model training. Rather, these patterns emerge from its architecture and training data. 
The patterns partially
align with IPA representations based on the anatomy of a human mouth, which Llama does not have. We suspect that instances which break these patterns, such as \textipa{/u/}, \textipa{/a/}, and \textipa{/I/}, emerge from the usage of these particular phonemes in context\footnote{These particular phonemes are often present in English language diphthongs, which could explain the divergence from anatomical characteristics.}.

As an attempt to show this portion of Llama's internal vowel model in a human-readable manner, we construct a variant of the IPA vowel chart which is consistent with our findings, populated with common English vowel sounds (Figure \ref{fig:ipa_chart}). This chart maintains the surprising emergent colinearities we find in our PCA of the vowel vectors, as shown by the overlay of this chart in Figure \ref{fig:new_ipa_over_pca}. H13L12 result vectors also align with these geometries (see Appendix \ref{resvec_clusters}).


We note that the representations in Figure \ref{fig:ipa_chart} do not necessarily constitute a comprehensive account of Llama's internal representations of phonetic information. However, the fact that we see some degree of linearity --- and the fact that this linearity aligns, to some degree, with existing anatomical representations --- indicates the presence of some kind of relativistic world model of vowel phonemes between the embedding space and the result vector of H13L12.

\section{Future Work}
While our experiments provide evidence of robust phonetic representation in \verb|Llama-3.2-1B-Instruct|'s embedding and residual stream, several important questions remain. Our findings open up multiple avenues for further research, especially regarding the mechanisms and structures through which 
the model propagates and processes phonetic information.
In particular, we are interested in further characterizing
H13L12's behaviors,
better understanding the cross-lingual features in the model's phonetic system, and isolating 
the propagation of phonetic information.

\section{Conclusion}


Our investigation provides strong evidence that \verb|Llama-3.2-1B-Instruct| contains a rich internal phonetic model which partially diverges from human anatomical phonetic models. We demonstrate the recoverability of phonetic information from token embeddings, which produces directions for each phoneme in latent space, and we identify a ``phoneme mover head" (H13L12) that moves and promotes phonetic information during rhyming tasks.
We use the result vectors of this head to fit a phonetically-informed PCA. Applying this PCA to our latent space phoneme vectors yields a geometric arrangement of phonemes. The pattern of vowel phoneme vectors in particular suggests the use of an internal vowel model which partly aligns with anatomically-informed vowel taxonomies.



\pagebreak
\bibliography{custom}

\appendix

\section{Emergence of third-party vowels in interventions}

\begin{figure}[htb]
    \centering
    \includegraphics[width=1.0\linewidth]{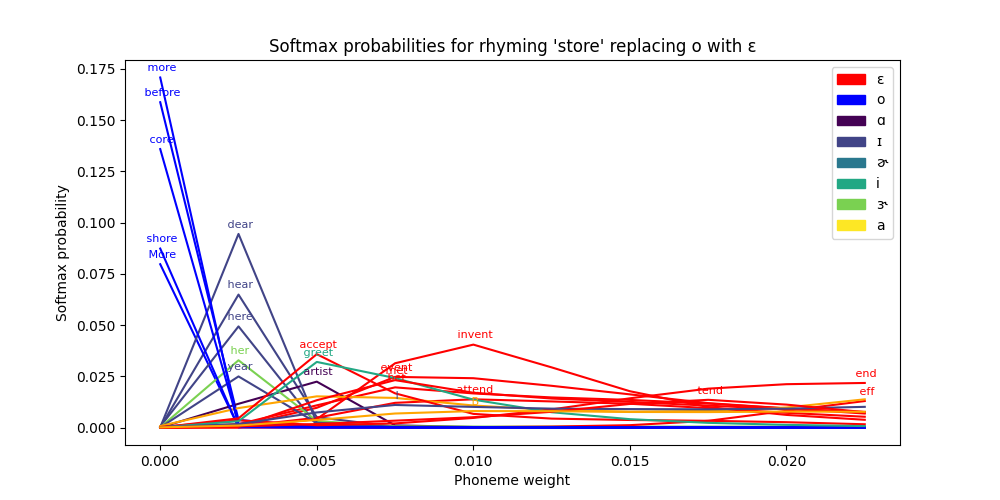}
    \caption{Example intervention on rhymes with \texttt{store}. When replacing the \textipa{/o/} sound with \textipa{/E/} at small phoneme weight values, the model predicts words containing \textipa{/i/}. This is consistent across multiple starting words containing \textipa{/o/}.}
    \label{fig:store}
\end{figure}

In the Figure \ref{fig:leet} example, the vowel sounds remain relatively isolated.
Sometimes, 
however,
unexpected third-party vowel sounds appear on the fringes, such as small \textipa{/o/}-to-\textipa{/E/} interventions manifesting \textipa{/i/} sounds, as shown in Figure \ref{fig:store}.

These ``intermediate" vowel sounds often appear in consistent patterns for consistent $\xi$ and $\mu$ vowels, regardless of the consonants involved in the starting word, which suggests that the model internally represents vowel sounds in the embedding space with meaningful relationships to each other. This motivates our attention to colinearities of phoneme vectors in PCA space in Section \ref{geometry}. When applying this methodology to non-phonetic interpretability tasks, we suggest carefully studying the presence of third-party characteristics (i.e. attributes of model results which are not strictly related to the specific intervention) in order to inform both the choice of dimensionality reduction technique and the expected emergent geometric patterns in PCA space.

\section{Activation patching results} \label{patchingres}
\begin{figure}[htb]
    \centering
    \includegraphics[width=1.0\linewidth]{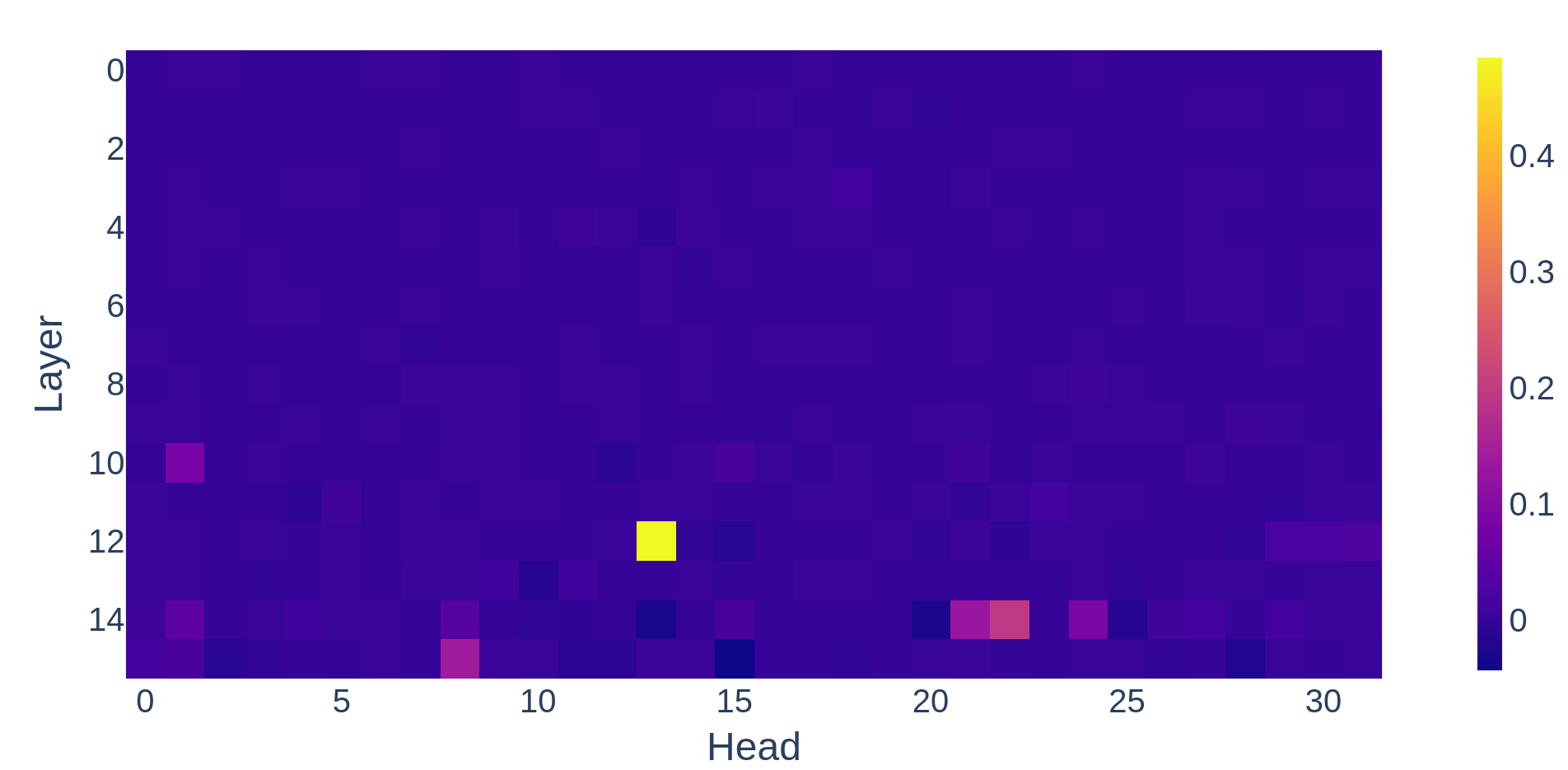}
    \caption{The mean normalized logit differences for our activation patching experiment described in Section \ref{patching_setup}. Head 13 Layer 12 clearly has the highest value at 0.48.}\label{fig:patch_results}
\end{figure}

\section{Exploring the head dimension} \label{Exploring}
After discovering H13L12, we were interested in understanding both how the internal head dimension is organized (e.g. are there phoneme specific neurons in the output matrix?) and how phonetic information is pulled from the target word into the head.

\begin{figure}[htb]
    \centering
    \includegraphics[width=1.0\linewidth]{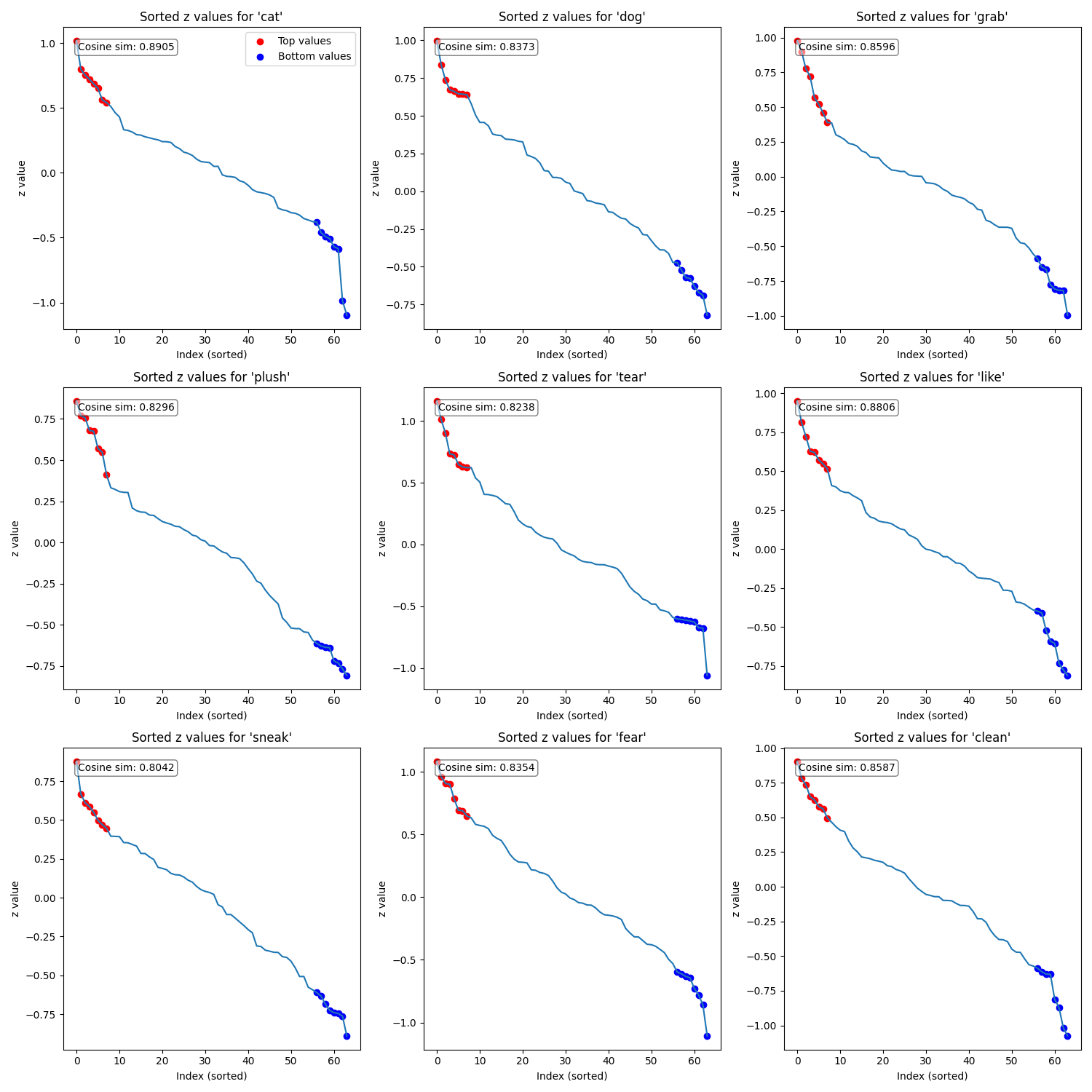}
    \caption{Sorted Z values with the top and bottom eight highlighted. Cosine similarity is between the full result vector and the vector generated by only taking the top and bottom eight values (setting every other index to zero) and passing it through the output matrix.}
    \label{fig:z-vals}
\end{figure}

The model seems to use complex interference patterns to approximate different phonetic directions. This is evidenced by the relative uninterpretability of the head dimension and the result matrix combined with the mutual consistency of similar phonemes within the head dimension with respect to cosine similarity. Notably, we were able to recover much of the result vector (measuring with cosine similarity) with the top and bottom (In terms of magnitude) eight values from the head vector, suggesting a kind of sparsity in the head dimension (See Figure \ref{fig:z-vals}). Interestingly, we found that all sixty four head dimensions were, for at least one word, present within the top or bottom eight z values. Future work could investigate the subspaces read from and written to by this head using more involved methods, like sparse coding \citep{kissane2024interpreting} or communication channels \citep{elhage2021mathematical, merullo2024talking}.

\section{Manual inspection of decoded result vectors} \label{resvec}
To better understand the role the phoneme mover head plays in downstream rhyme production, we perform a manual survey of one hundred different runs of our rhyme task (described in Section \ref{Probing_Embedding}), chosen randomly after filtering for tokenization from the Oxford 5000 dataset \cite{GÜZEL_2025}.

We judge result vectors according to the following criteria: we call a result vector (R.V.) ``\textit{coherent}" if and only if five of the top ten tokens it promotes contain similar phonemes to the target rhyme. For example, if the target word was \verb|plush| then the target rhyme would be \textipa{/2S/}. We consider the task to be passed if a correct rhyme is within the top ten tokens the model predicts as the next token.
\begin{table}[h]
    \centering
    \begin{tabular}{|c|c|c|}
        \hline
        & Coherent R.V. & Incoherent R.V. \\ \hline
        Pass & 55\% & 0\% \\ \hline
        Fail & 25\% & 18\% \\ \hline
    \end{tabular}
    \label{tab:result_vector}
\end{table}
Although our sample size is quite small, the effect is clear. Rarely occur, if ever, is the task completed successfully when an incoherent result vector is produced. In every example we analyzed, having a coherent R.V. was a prerequisite for completing the task.


\section{Result vectors cluster around the proposed vowel chart}
\label{resvec_clusters}

\begin{figure}
    \centering
    \includegraphics[width=0.8\linewidth]{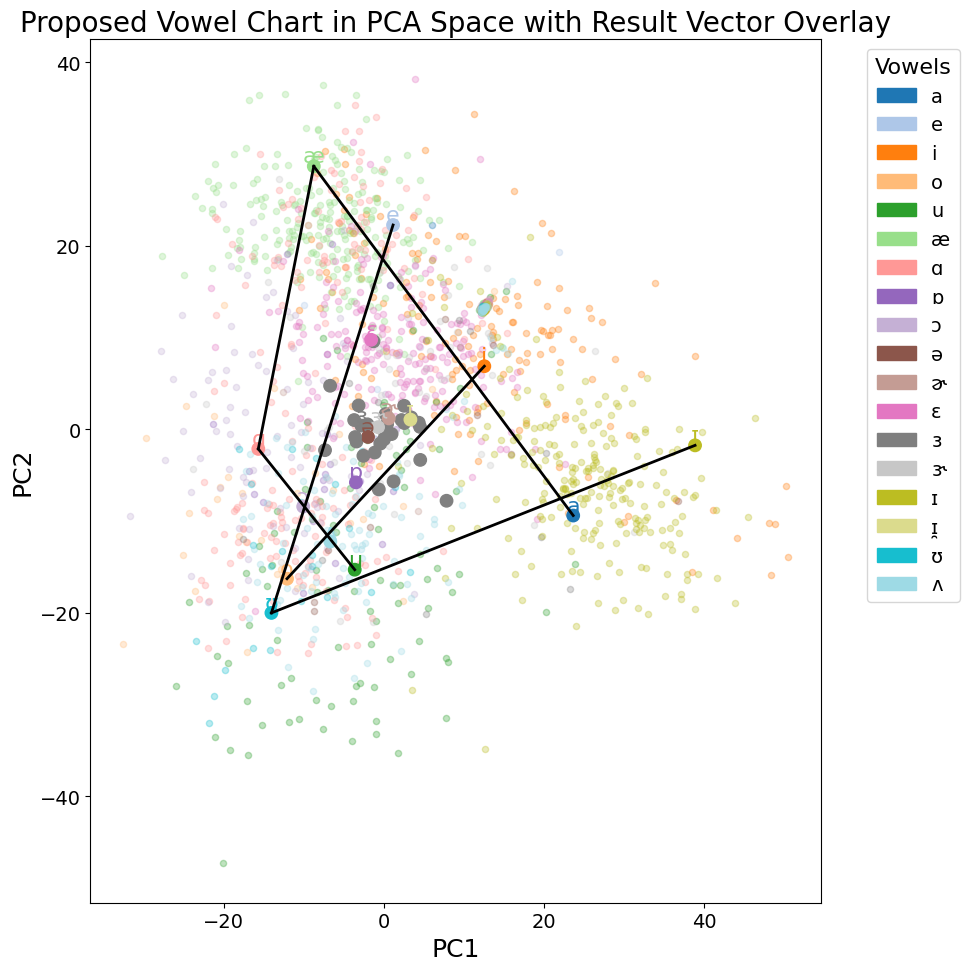}
    \caption{(High opacity) phoneme vectors corresponding to common English vowels in PCA space.\\
    (Low opacity) H13L12 result vectors of single-token single-vowel words in the context of our rhyming task, scaled and shifted. The relative geometry of these result vector clusters matches the geometry of the proposed vowel chart in Figure \ref{fig:ipa_chart}.}
    \label{fig:vowel-chart-resvec-overlay}
\end{figure}

The vowel chart we propose in Figure \ref{fig:ipa_chart} aligns with phoneme vector geometries in PCA space as shown in Figure \ref{fig:new_ipa_over_pca}. Importantly, this geometry is also present among result vectors. We use the same PCA to transform the H13L12 result vectors from 2000 single-token single-vowel words. Figure \ref{fig:vowel-chart-resvec-overlay} shows that up to linear transformation\footnote{The result vectors tend to have much smaller magnitude than our phoneme vectors due to differences in normalization, so we scale up the PCs of the result vectors by a factor of 25 and shift each component by +8.}, result vectors cluster around the phoneme vectors corresponding to their vowels.

\end{document}